\documentclass[journal]{IEEEtran}
\ifCLASSINFOpdf
   \usepackage[pdftex]{graphicx}
\else
   \usepackage[dvips]{graphicx}
\fi

\usepackage[cmex10]{amsmath}

\usepackage[tight,footnotesize]{subfigure}

\usepackage{url}

\newcommand{\comment}[1]{}

\begin{document}
%
\title{Efficient and Accurate Gaussian Image Filtering Using Running Sums}
%
%
%

\author{Elhanan~Elboher 
        and~Michael~Werman
\thanks{E. Elboher and M. Werman are with the Hebrew University of Jerusalem, Israel.}

}

\maketitle

\begin{abstract}
This paper presents a simple and efficient method to convolve an image with a
Gaussian kernel.
The computation is performed in a constant number of operations per pixel
using running sums along the image rows and columns.
We investigate the error function used for kernel approximation
and its relation to the properties of the input signal.
Based on natural image statistics we propose a quadratic form kernel error function
so that the output image $l_2$ error is minimized.
We apply the proposed approach to approximate the Gaussian kernel 
by linear combination of constant functions.
This results in very efficient Gaussian filtering method.
Our experiments show that the proposed technique is faster than state of the art
methods while preserving a similar accuracy.
\end{abstract}

\begin{IEEEkeywords}
Non uniform filtering, Gaussian kernel, integral images, natural image statistics.
\end{IEEEkeywords}

%
\IEEEpeerreviewmaketitle

\section{Introduction}
\label{sec:intro}

\IEEEPARstart{I}{mage} filtering is an ubiquitous image processing tool, 
which requires fast and efficient computation.
When the kernel size increases,
direct computation of the kernel response requires more operations and 
the process becomes slow.

Various methods have been suggested for fast convolution with specific kernels
in linear time (see related work in Section~\ref{sec:related}).
An important kernel is the Gaussian, which is used in many applications.

In this paper we present an efficient filtering algorithm for separable non uniform kernels
and apply it for very fast and accurate Gaussian filtering.
Our method is based on one dimensional \emph{running sums} 
(\emph{integral images}) along single rows and columns of the image.
The proposed algorithm is very simple and can be written in a few lines of code.
Complexity analysis, as well as experimental results show that it is faster than
state of the art methods for Gaussian convolution while preserving similar
approximation accuracy. 

An additional contribution of this paper is an analysis of the relation between
the kernel approximation and the approximation of the final output, the filtered image.
Usually, the approximation quality is measured in terms of the 
difference between the kernel and its approximation.
However, minimizing the kernel error does not necessarily minimize the 
error on the resulting image.
To minimize this error, natural image statistics should be considered.

In Section~\ref{sec:approx} we investigate 
the kernel approximation error 
function. 
Based on natural image statistics we find a quadratic form
kernel error measurement which minimizes
the $l_2$ error on the output pixel values.

The next section reviews related methods for fast non uniform image filtering.
Section~\ref{sec:alg} presents the filtering algorithm and discusses its computational aspects.
Section~\ref{sec:approx} discusses the relation between kernel approximation and natural image statistics.
Section~\ref{sec:res} presents our experimental results.
We conclude in Section~\ref{sec:conc}.

\section{Related Work}
\label{sec:related}

In the following we review the main approaches to accelerate image filtering.
We describe in more detail the integral image based approach on which the current work is based.

{\bf General and Multiscale Approaches.}
Convolving an image of $N$ pixels with an arbitrary kernel of size $K$
can be computed directly in $O(NK)$ operations,
or using the Fast Fourier Transform in $O(N \log K)$ operations~\cite{gonzalez1992woods}.

A more efficient approach is the linear time multiscale computation using
image pyramids~\cite{burt1981fast, burt1983laplacian}.
The coarser image levels are filtered with small kernels and the results are interpolated
into the finer levels. This approximates the convolution of the image
with a large Gaussian kernel.

{\bf Recursive Filtering.}
The recursive method is a very efficient filtering scheme for one dimensional 
(or separable) kernels. The infinite impulse response (IIR) of the desired kernel
is expressed as a ratio between two polynomials in Z space~\cite{proakis1996digital}. 
Then the convolution with a given signal is computed by difference equations.
Recursive algorithms were proposed for approximate filtering with
Gaussian kernels~\cite{deriche1987using,deriche1990fast,deriche1993recursively,young1995recursive,van1998recursive,jin1997recursive},
anisotropic Gaussians~\cite{geusebroek2003fast}
and Gabor kernels~\cite{young2002recursive}.

The methods of Deriche~\cite{deriche1987using,deriche1990fast,deriche1993recursively} 
and Young and van Vliet~\cite{young1995recursive,van1998recursive}
are the current state of the art for fast approximate Gaussian filtering.
Both methods perform two passes in opposite directions, in order to consider the
kernel response both of the forward and backward neighbors.
The method of Young and van Vliet's requires less arithmetic operations per pixel.
However, unlike Deriche's method the two passes cannot be parallelized.

Tan et al.~\cite{tan2003performance} evaluated the performance of both methods
for small standard deviations using normalized RMS error.
While Deriche's impulse response is more accurate,
Young and van Vliet performed slightly better on a random noise image.
No natural images were examined.
Section~\ref{sec:res} provides further evaluation of these methods.

{\bf Integral Image Based Methods.}
Incremental methods such as \emph{box filtering}~\cite{mcdonnell1981box}
and \emph{summed area tables}, known also as 
\emph{integral images}~\cite{crow1984summed,viola2001rapid},
cumulate the sum of pixel values along the image rows
and columns. In this way the sum of a rectangular region can be computed using 
$O(1)$ operations independent of its size.
This makes it possible to convolve an image very fast with uniform kernels.

Heckbert~\cite{heckbert1986filtering} 
generalized integral images for polynomial kernels
of degree $d$ using $d$ repeated integrations.
Derpanis et al.~\cite{derpanis2007fast} applied this scheme for a polynomial 
approximation of the Gaussian kernel.
Werman~\cite{werman2003fast} introduced another generalization
for kernels which satisfy a linear homogeneous equation (LHE).
This scheme requires $d$ repeated integrations, where $d$ is the LHE order.

Hussein et al.~\cite{hussein2008kernel} proposed Kernel Integral Images (KII)
for non uniform filtering.
The required kernel is expressed
as a linear combination of simple functions.
The convolution with each such functions is computed separately using integral images.
To demonstrate their approach, the authors approximated the Gaussian kernel by a linear 
combination of polynomial functions based on the Euler expansion.
Similar filtering schemes were suggested by Marimon~\cite{marimon2010fast}
who used a combination of linear functions to form pyramid shaped kernels
and by Elboher and Werman~\cite{elboher2011cosine} who used a combination of
cosine functions to approximate the Gaussian and Gabor kernels and 
the bilateral filter~\cite{tomasi2002bilateral}.

{\bf Stacked Integral Images.}
The most relevant method to this work
is the Stacked Integral Images (SII) proposed by Bhatia et al.~\cite{bhatia2010stacked}.
The authors approximate non uniform kernels by a 'stack' of box filters,
i.e. constant 2D rectangles, which are all computed from a \emph{single} integral image.
The simplicity of the used function and not using multiple integral images 
makes the filtering very efficient.

The authors of SII demonstrated their method for Gaussian smoothing.
However, they approximated only specific 2D kernels,
and found for each of them a local minima of a non-convex optimization problem.
Although the resulting approximations can be rescaled, they are not very accurate 
(see Section~\ref{sec:res}).
Moreover, the SII framework does not exploit the separability of the Gaussian kernel.

\begin{figure}[t]
  \centering

  \subfigure[Gaussian kernel.] 
  {
  \includegraphics[width=0.45\linewidth]{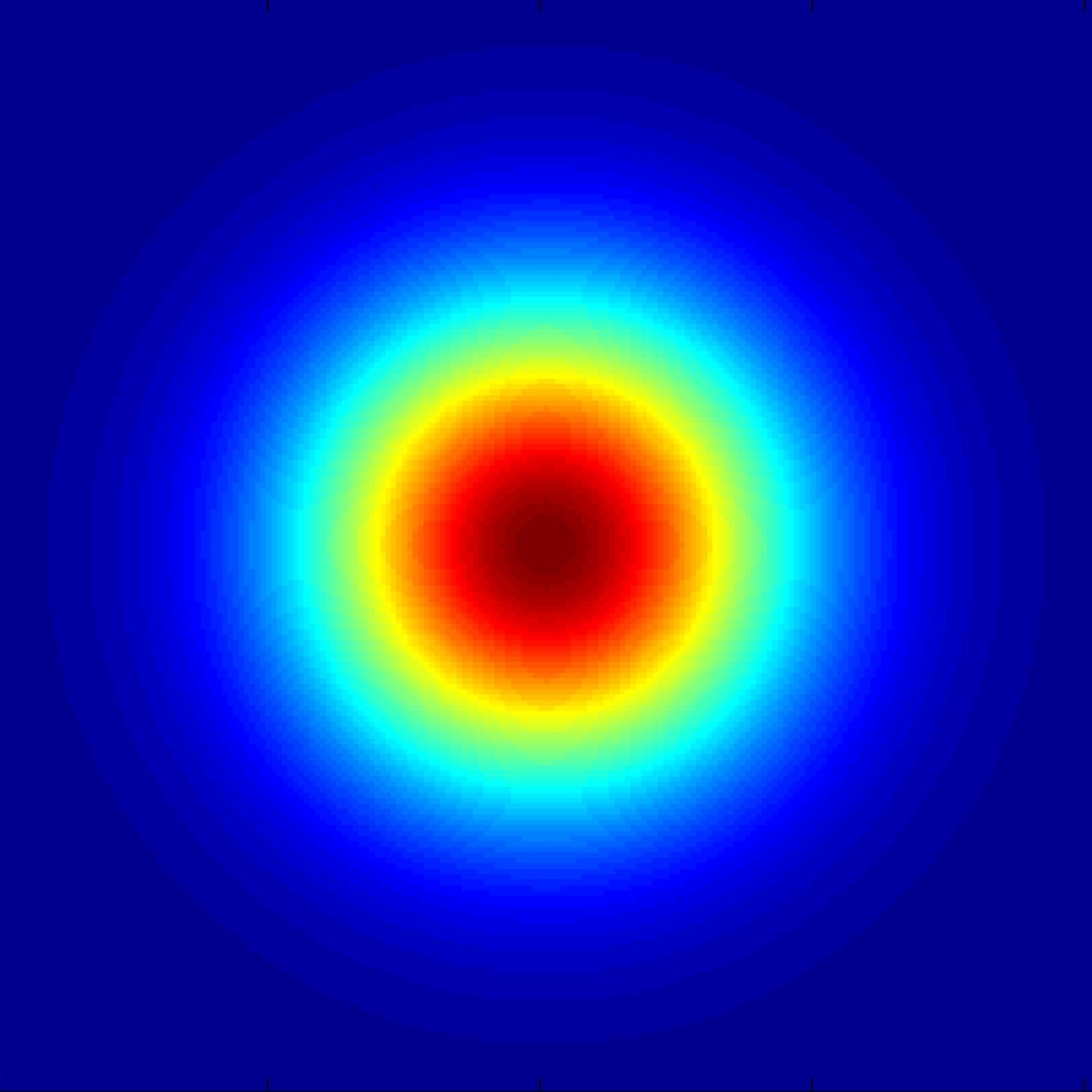}
  }
  \subfigure[SII~\cite{bhatia2010stacked} approximation.] 
  {
  \includegraphics[width=0.45\linewidth]{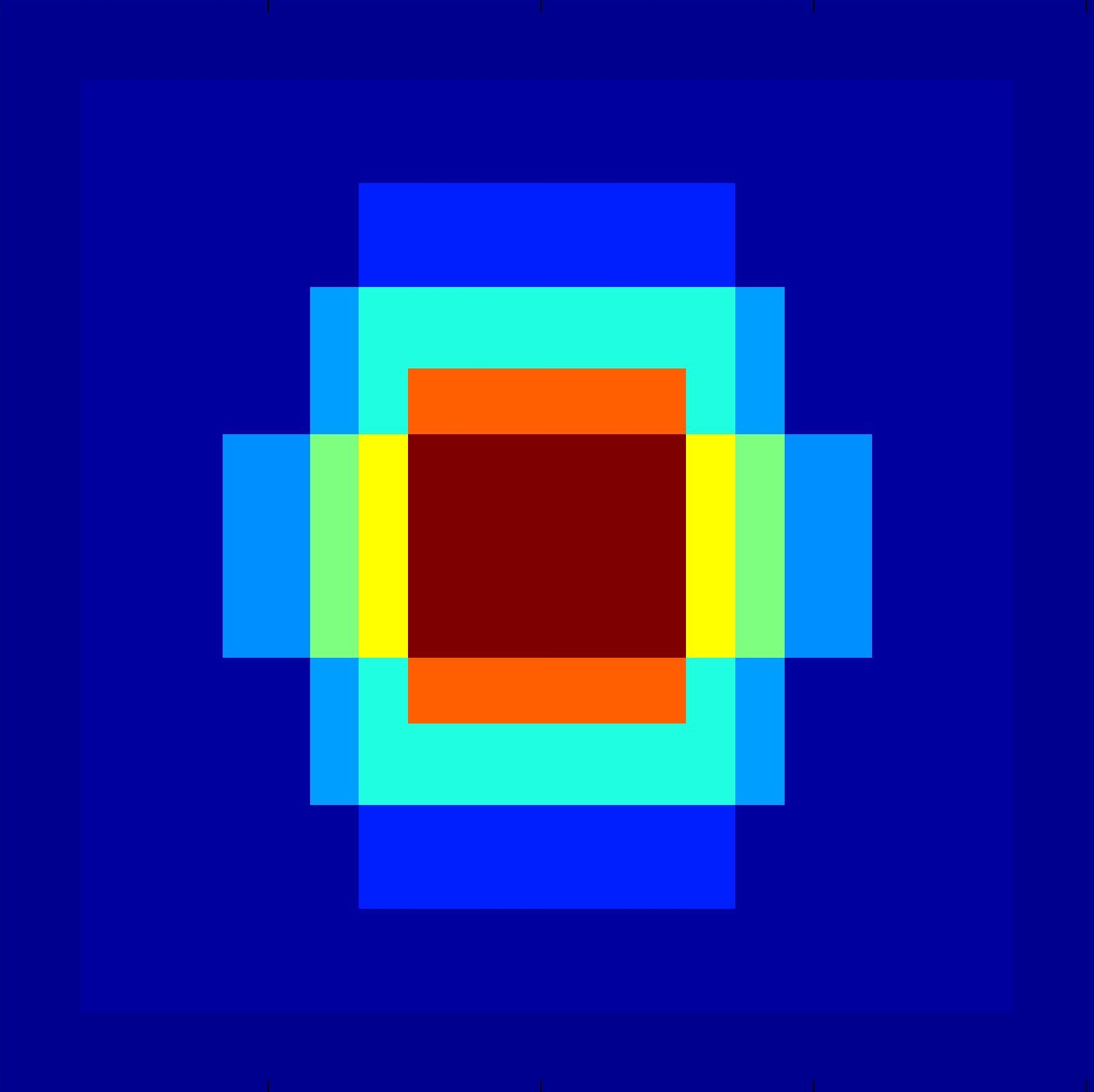}
  }

  \subfigure[Proposed approximation ($4$ constants).] 
  {
    \includegraphics[width=0.45\linewidth]{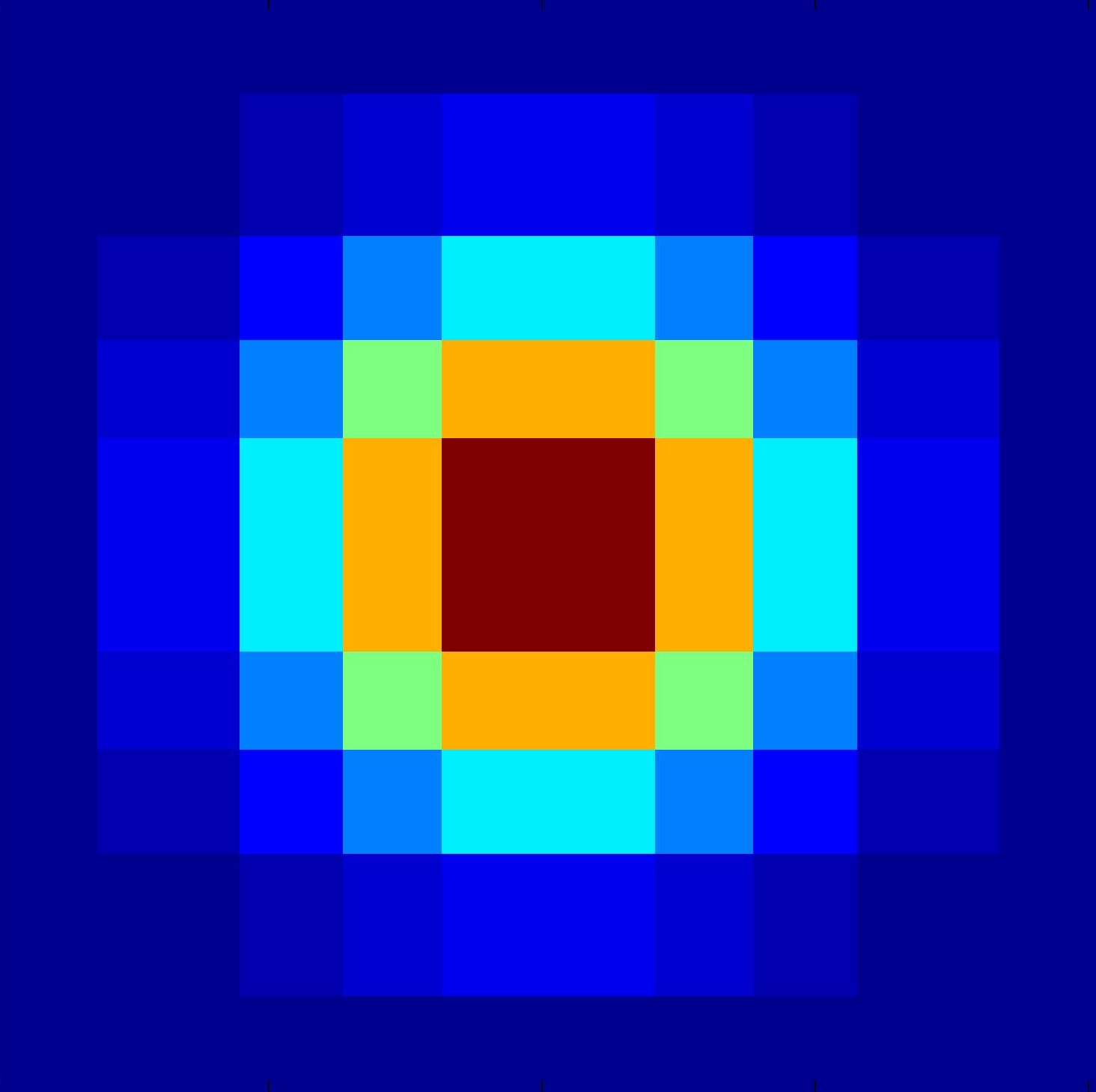}
  }
  \subfigure[Proposed approximation ($5$ constants).] 
  {
  \includegraphics[width=0.45\linewidth]{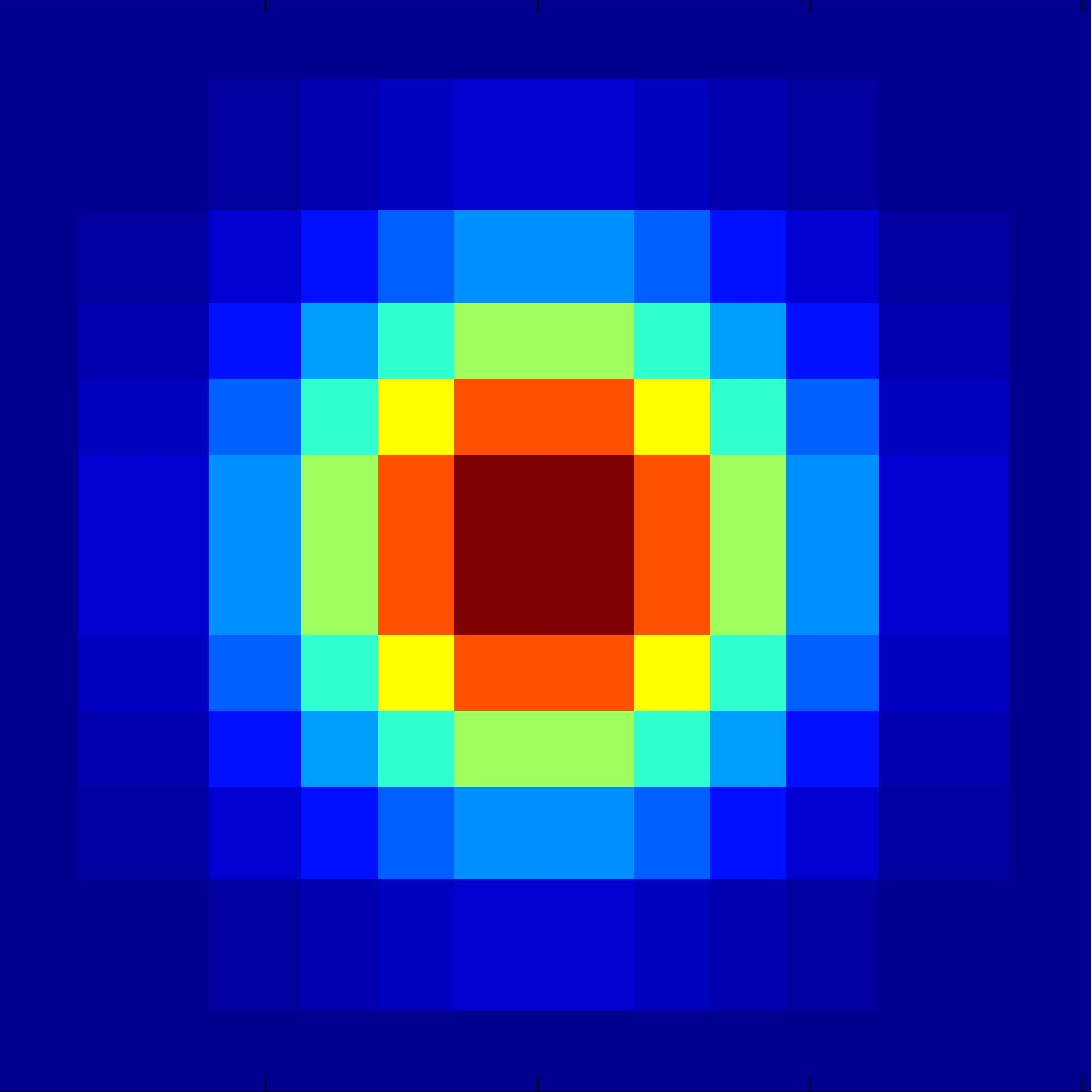}
  }
  
  \caption{Comparison of (a) exact Gaussian kernel, 
    (b) Stacked Integral Images~\cite{bhatia2010stacked} with $5$ 2D boxes,
    and the proposed method with $4$ constants (c) and $5$ constants (d).
    Our proposed approximation is richer and more accurate since it utilizes 
    the Gaussian separability. Instead of using 2D boxes, we use 1D segments
    to filter the rows and then the columns.}
  \label{fig:pulse}
\end{figure}

As shown in this paper, utilizing the separability property 
we find an optimal kernel approximation which can be scaled to
any standard deviation.
As shown in Figure~\ref{fig:pulse}, this approximation is richer and more accurate.
Actually, separable filtering of the row and the columns by $k$ one dimensional constants
is equivalent to filtering by $2k-1$ two dimensional boxes.
The separability property can also be used for an efficient computation 
both in time and space, as described in Section~\ref{sec:alg}.

\begin{figure}[t]
  \centering
  \includegraphics[width=0.8\linewidth]{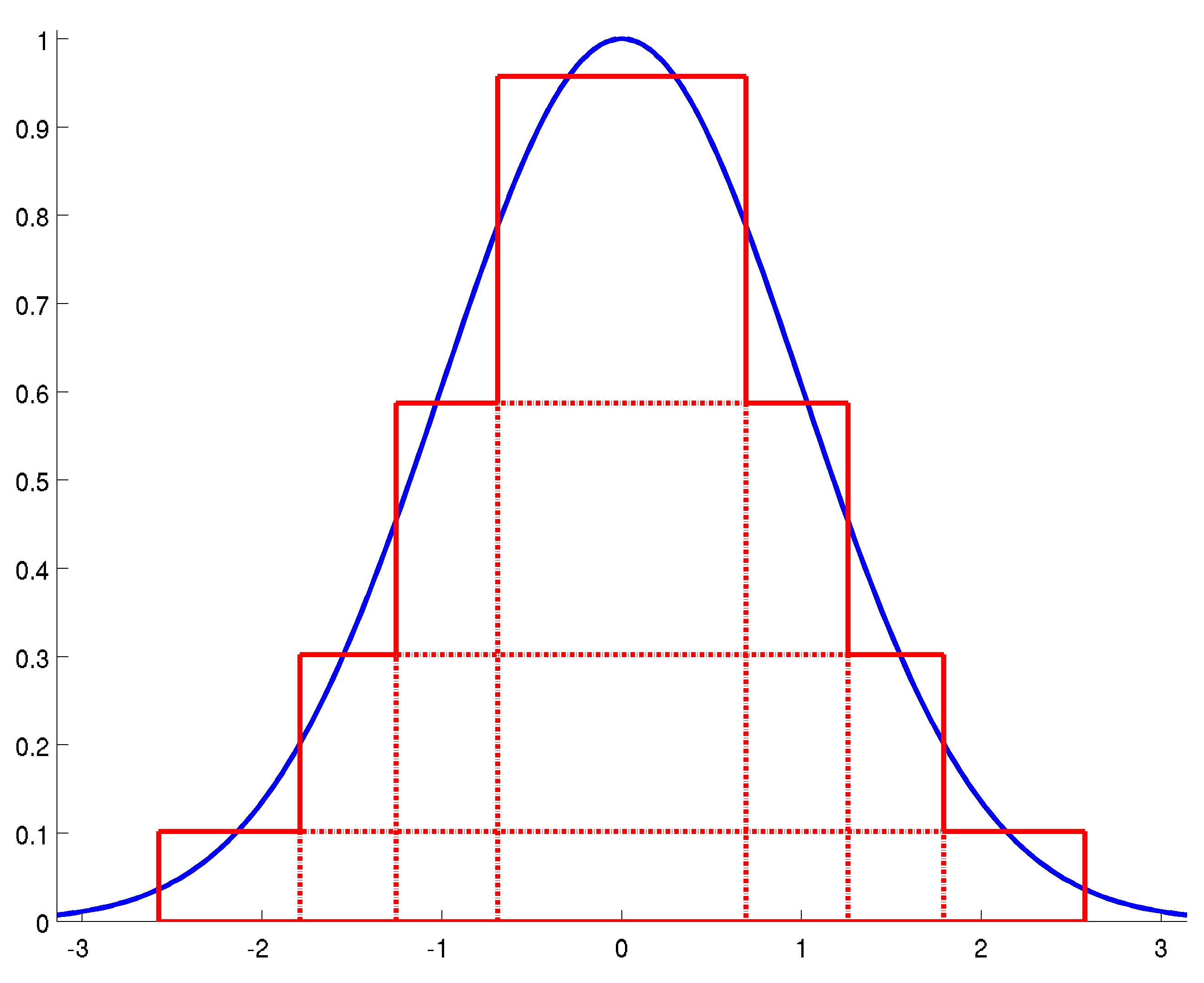}
  \caption{Approximation of the 1D Gaussian kernel on $[-\pi,\pi]$ 
    by $k=4$ piecewise constant functions.
    The dashed lines show the equivalence between $4$ constant 'slices' and $7$
    constant 'segments' (Equation~\ref{eq:ciwi}).}
  \label{fig:1Dapprox}
\end{figure}

\section{Algorithm}
\label{sec:alg}

\subsection{Piecewise Constant Kernel Decomposition}
\label{sec:kerdec}

Consider the convolution $f*K$ of a function $f$ with a kernel $K$.
For simplicity we first discuss the case in which $f$ and $K$ are one dimensional.
In the following (Section~\ref{subsec:higher}) we generalize the discussion to 
higher dimensions.

Suppose that the support of the kernel $K$ is $r$, i.e. that $K$ is zero outside of $[0,r]$.
Assume also that we are given a partition $P = (p_0,p_1,...p_k)$,
in which $0 \leq p_0 < p_1 < p_2$ $...$ $< p_{k-1} < p_k \leq r$.
Thus, the kernel $K$ can be approximated by a linear combination
of $k$ simple functions $K_i$, $i = 1...k$:

\begin{equation}
\label{eq:k_i}
K_i(t) = 
 \left\{ \begin{array}{rl}
 c_i &\mbox{ if $p_{i-1} \leq t <p_i$} \\
  0 &\mbox{ otherwise}
       \end{array} \right.
\end{equation}

\noindent
Using the above approximation
\begin{equation}
\label{eq:kerdec-1}
f * K \approx \sum_{i=1}^{k} f * K_i
\end{equation}

\noindent
which can be computed very efficiently, as described in Section~\ref{subsec:algo}.

\subsection{Symmetric Kernels}
\label{subsec:sym}

Consider the case in which $K$ is a \emph{symmetric} kernel on $[-r,r]$.
The approximation can be limited to the range $[0,r]$.
Each constant $c_i$ can be used both in the negative interval 
$[-p_i,-p_{i-1}]$ and in the positive one $[p_{i-1},p_i]$,
however, this requires $2k$ constant function.
Actually, the same approximation can be computed using 'weighted slices':

\begin{equation}
\label{eq:S_i}
S_i(t) = 
 \left\{ \begin{array}{rl}
 w_i &\mbox{ if $-p_i < t < p_i$} \\
  0 &\mbox{ otherwise}
       \end{array} \right.
\end{equation}

\noindent
where $w_i = c_i-c_{i-1}$.
Now the kernel $K$ is approximated by sum of $k$ constant functions which are non zero
in the overlapping intervals $[-p_i,p_i]$,
with weights $w_i = c_i-c_{i-1}$. 
The approximation of $K$ remains the same:


\begin{equation}
  \label{eq:ciwi}
  \begin{array}{lll}
   \displaystyle{ \sum_{i=1}^k{S_i(t)}}   
&
= \displaystyle{\sum_{i:-p_i < t < p_i}{w_i}} 
&  
= \displaystyle{\sum_{i:-p_i < t < p_i}{(c_i - c_{i-1})}} 
\\
&&\\
& 
= \displaystyle{\sum_{i:p_{i-1} \leq |t| < p_i}{c_i}} 
&
=~~~~\displaystyle{\sum_{i=1}^k K_i(t)} 
  \end{array}
\end{equation}



\noindent
This equality is illustrated in Figure~\ref{fig:1Dapprox}.

\subsection{1D Filtering Algorithm}
\label{subsec:algo}


In the following we describe the algorithm for the case of a
\emph{symmetric} kernel $K$ (Section~\ref{subsec:sym}).

Given a 1D discrete function $f(x)$ and $k$ piecewise constant functions $S_i$,
we compute the approximated convolution (Equation~\ref{eq:kerdec-1}) as follows:

\begin{enumerate}

\item 
Compute the cumulative sum of $f(x)$,
\begin{equation}
\label{eq:alg1}
I(x) = \sum_{x'=0}^{x} f(x')
\end{equation}

\item
Compute the convolution result for each pixel $x$,
\begin{equation}
\label{eq:indep}
\sum_{i=1}^{k} w_i (I(x+p_i) - I(x-p_i-1))
\end{equation}

\end{enumerate}

The total cost of steps $1$ and $2$ is $2k$ additions and $k$ multiplications
per image pixel.
In the 1D case memory access is not an issue, as all the elements are usually
in the cache.

\begin{table}[t]

\begin{center}
\begin{tabular}{|c||c|c|}
\hline
Method                                        &  Additions  &  Multiplications  \\
\hline
\hline
Direct                            &    $h+w-2$            &  $h+w$    \\
\hline
FFT                               & $O(log(hw))$  &  $O(log(hw))$ \\
\hline
KII~\cite{hussein2008kernel}                  &   $53$      &  $18$      \\
\hline
CII~\cite{elboher2011cosine}                  &   $12k-8$   &  $8k-8$    \\
\hline
SII~\cite{bhatia2010stacked}                  &   $4k+1$    &   $k$    \\
\hline
Deriche~\cite{deriche1993recursively}         &   $8k-2$    &   $8k$   \\
\hline
Young and                           &   $4k$      &   $4k+4$ \\
van Vliet~\cite{young1995recursive} & & \\
\hline
Proposed method                               &   $4k$      &   $2k$   \\
\hline
\end{tabular}
\end{center}
\caption{Arithmetic operations per image pixel (Section~\ref{subsec:higher}).}
\label{tbl:opers}

\end{table}

\subsection {Higher Dimensions and More Computational Aspects}
\label{subsec:higher}

We now describe the case of convolving a 2D image $f(x,y)$ with a 
\emph{separable} 2D kernel $K$. 
A kernel $K$ is separable if it can be expressed as a convolution of 
1D filters $K_x * K_y^{T}$.
The convolution $f * K$ can be computed by first convolving the image rows with $K_x$
and then the columns of the intermediate result by $K_y$.
Hence, filtering an image with a separable 2D kernel only 
doubles the computational cost of the 1D case.

The space complexity is also very low.
Since the convolution of $K_x$ with each row (and also $K_y$ with each column) 
is independent, filtering an image of size $n \times m$ 
requires only $O(\max(n,m))$ additional space over the input and the output images
for storing the cumulative sum of a single row or column.

Similarly to the 2D case, the filtering scheme can be extended to $d$-dimensional
separable kernels using $2dk$ additions and $dk$ multiplications per single pixel.
The additional required space is $O(\max_i(n_i))$, where $n_i$ $(i = 1...d)$ are the
sizes of the signal dimensions.

Table~\ref{tbl:opers} counts the required operations per image pixel
for 2D image filtering by a $h$$\times$$w$ separable kernel.
The parameter $k$ denotes the number of terms (constants, 2D boxes etc.)
depending on the specific method.
Notice that experimentally 
 our proposed method with $3$ constants is more accurate  
than SII~\cite{bhatia2010stacked}
with $5$ boxes and has an  accuracy similar to 
the recursive methods with $3$ coefficients
(Section~\ref{sec:res}, Figure~\ref{fig:psnr}).
This means that our proposed method requires less operations than Deriche~\cite{deriche1993recursively}
and all the integral image based methods,
and a comparable number of operations to Young and van Vliet~\cite{young1995recursive}.

The proposed method is also convenient for parallel computation.
The summation step (Equation~\ref{eq:alg1}) can be parallelized as described
in Section~4.3 of~\cite{hendeby2007graphics}, 
while the next step (Equation~\ref{eq:indep}) computes the response of each pixel
independently of its neighbors.
On the other hand, the recursive methods can be parallelized only partially --
e.g. by filtering different rows independently.
However, within each row (or column) all the computations are strongly dependent.

Notice also that Equation~\ref{eq:indep} can be computed for 
a small percentage of the image pixels.
This can accelerate applications in which the kernel response is computed
for sampled windows such as image downscaling.
The recursive methods do not have this advantage, since they compute the
kernel response of each pixel using the responses of its neighbors.

\comment{
\begin{table*}[t]

\begin{center}
\begin{tabular}{|c|l|l|l|l|}
\hline
\#constants &   $l_2$ partition &  $l_2$ weights  ($c_i$) & autocorrelation           & autocorrelation  weights ($c_i$)  \\
  ($k$)     & indices ($p_i$)   &        & partition indices ($p_i$) &  \\
\hline
\hline
$k = 3$    &  24, 42, 64        & 0.9095, 0.5755, 0.2522 & 23, 46, 76               & 0.9495, 0.5502, 0.1618\\
\hline
$k = 4$    &  20, 34, 49, 64    & 0.9350, 0.6871, 0.4216, 0.1813  & 19, 37, 56, 82           & 0.9649, 0.6700, 0.3376, 0.0976\\
\hline
$k = 5$    &  18, 30, 41, 54, 73          & 0.9465, 0.7418, 0.5282, 0.3236, 0.1384  & 16, 30, 44, 61, 85 & 0.9738, 0.7596, 0.5031, 0.2534, 0.0739\\
\hline
\end{tabular}
\end{center}
\caption{Algorithm parameters.}
\label{tbl:params}

\end{table*}
}

\section{Kernel Approximation}
\label{sec:approx}

The approximation of a one dimensional kernel $K(t)$ by $k$ constant functions
is determined by the partition $P$ and the constants $c_i$.
Finding the best parameters is done by minimizing an approximation error function
on the desired kernel.
Indeed, our real purpose is to minimize the error on the \emph{output image}, 
which is not necessarily equivalent.
Section~\ref{sec:l2} relates the kernel approximation error and the output image error
using natural image statistics.
We define a quadratic form error function for the kernel approximation,
so that the output image $l_2$ error is minimized.
This error function is used in Section~\ref{sec:gapprox} to approximate the Gaussian kernel.

\subsection{Minimal Output $l_2$ Error}
\label{sec:l2}

We denote the input signal by $x$, the output signal by $y$,
and the kernel weights by $w$. 
The approximated output and kernel weights are denoted by $\hat{y}$ and $\hat{w}$ respectively.
The upper case letter $X$ denotes the Fourier transforms of the input $x$.

The squared $l_2$ error of the final result is given by

\begin{equation}
\label{eq:e2}
E_2 = \sum_i (y_i - \hat{y}_i)^2
\end{equation}

\noindent
Since $y_i = \sum_j{w_j x_{i+j}}$, $E_2$ can be expressed in terms of the 
input signal $x$ and the kernels $w,\hat{w}$:

\begin{equation}
  \label{eq:e2}
  \begin{array}{ll}
E_2 &= \displaystyle{\sum_i \Bigl( \sum_j{w_j x_{i+j}} - \sum_j{ \hat{w}_j x_{i+j}} \Bigr)^2} \\
&\\
&= \displaystyle{\sum_{i,j,k} (w_j - \hat{w}_j) (w_k - \hat{w}_k)  x_{i+j} x_{i+k}} \\
&\\
&= \displaystyle{\sum_{j,k} \Bigl( (w_j - \hat{w}_j) (w_k - \hat{w}_k)  \sum_i{x_{i+j} x_{i+k}} \Bigr)} \\
\end{array}
\end{equation}

\comment{
\begin{equation}
E_2 = \sum_i \left( \sum_j{w_j x_{i+j}} - \sum_j{ \hat{w}_j x_{i+j}} \right)^2
\end{equation}

\[
= \sum_{i,j,k} (w_j - \hat{w}_j) (w_k - \hat{w}_k)  x_{i+j} x_{i+k} 
\]

\[
= \sum_{j,k} \left( (w_j - \hat{w}_j) (w_k - \hat{w}_k)  \sum_i{x_{i+j} x_{i+k}} \right)
\]
}

\noindent
The only term which involves the input signal $x$ is the inner sum, which we denote as
$A_{jk} = \sum_i{x_{i+j} x_{i+k}}$. 

Note that $A_{jk}$ is the the autocorrelation of $x$ at location $j$$-$$k$.
The error can be therefore expressed as

\begin{equation}
\label{eq:mahal}
E_2 =  (w - \hat{w})^{\texttt{T}} A (w - \hat{w})
\end{equation}

\noindent
where $A$'s entries are given by the autocorrelation of the signal $x$.

In order to make $E_2$ independent of the values of a specific $x$, 
we make use of a fundamental property of natural images introduced by Field~\cite{field1987relations}
 the Fourier spectrum of a natural image in each frequency $u$
is proportional to $\frac{1}{u}$,

\begin{equation}
|X_u| \propto \frac{1}{u}
\end{equation}

\noindent
Applying the Convolution Theorem,  the autocorrelation of $x$ is

\begin{equation}
|X_u|^2 \propto \frac{1}{u^2}
\end{equation}

\noindent
Hence, $A_{jk}$ should be proportional to $\Phi$, the inverse Fourier transform
of $\frac{1}{u^2}$.

Of course, the definition of $\Phi$ is problematic since 
$\frac{1}{u^2}$ is not defined for $u=0$.
Completing the missing value by an arbitrary number affects all $\Phi$ values.
However,  additional information is available which helps to 
complete the missing value.

Assuming that the values of $x$  are uniformly distributed in $[0,1]$, we  find 
that $\Phi_{_0}$ (which equals to $A_{jj}$, the correlation of a pixel with itself) 
is proportional to $\int_0^1{ x^2 dx } = \frac{1}{3}$.
In addition, assuming that distant pixels are uncorrelated, the boundary values $\Phi_{_r}, \Phi_{_{-r}}$
are proportional to $\mu_x^2 = \frac{1}{4}$.
This means that the ratio $\frac{\Phi_{_0}}{\Phi_{_r}}$ should be $\frac{4}{3}$.
Determining the missing zero value in the Fourier domain as $16.5$ results in
such a function $\Phi$.

\begin{figure*}[t]
  \centering

  \subfigure[Speedup] 
  {
  \includegraphics[width=0.7\linewidth]{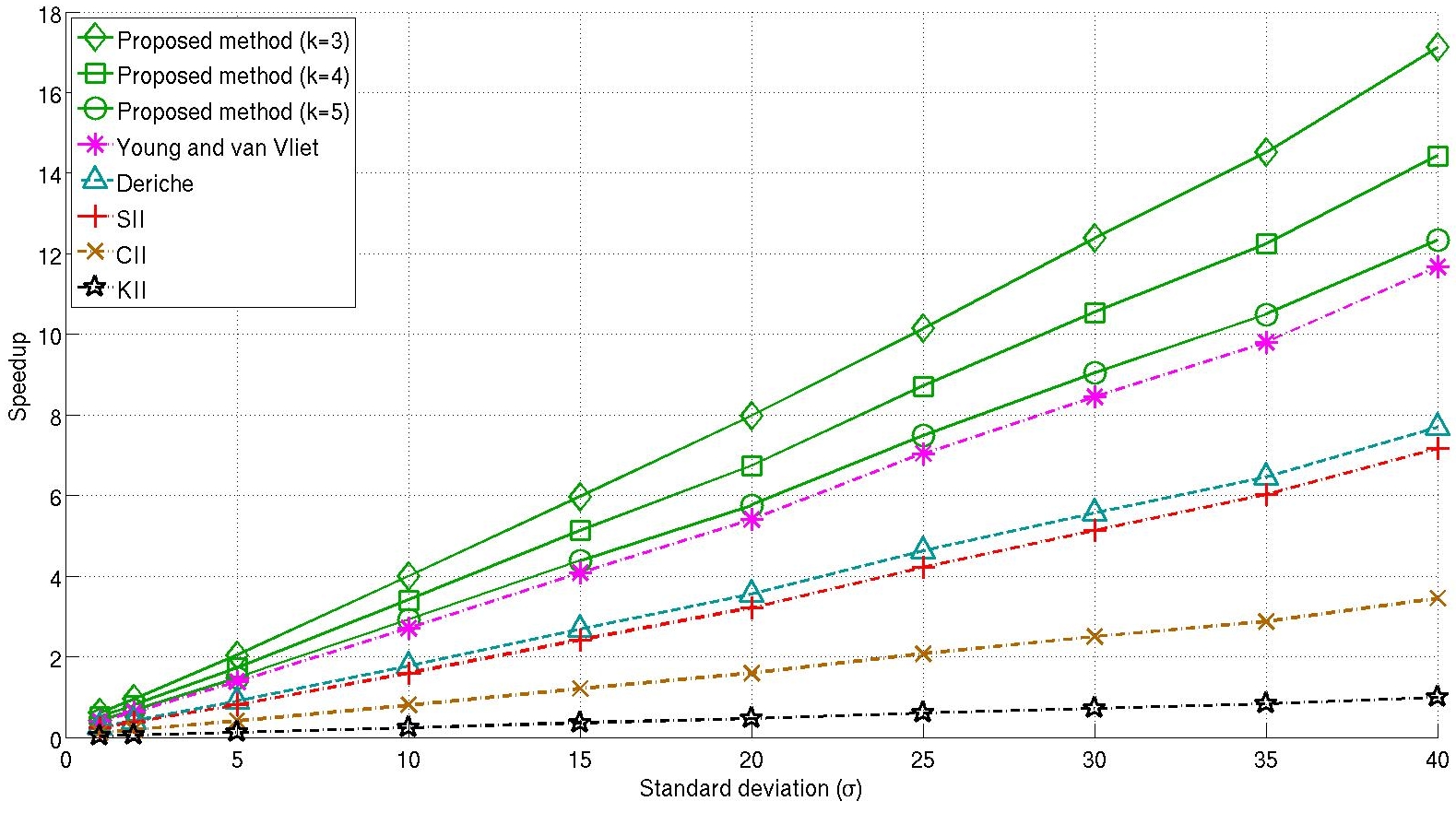}
  \label{fig:speedup}
  }

  \subfigure[Accuracy] 
  {
  \includegraphics[width=0.7\linewidth]{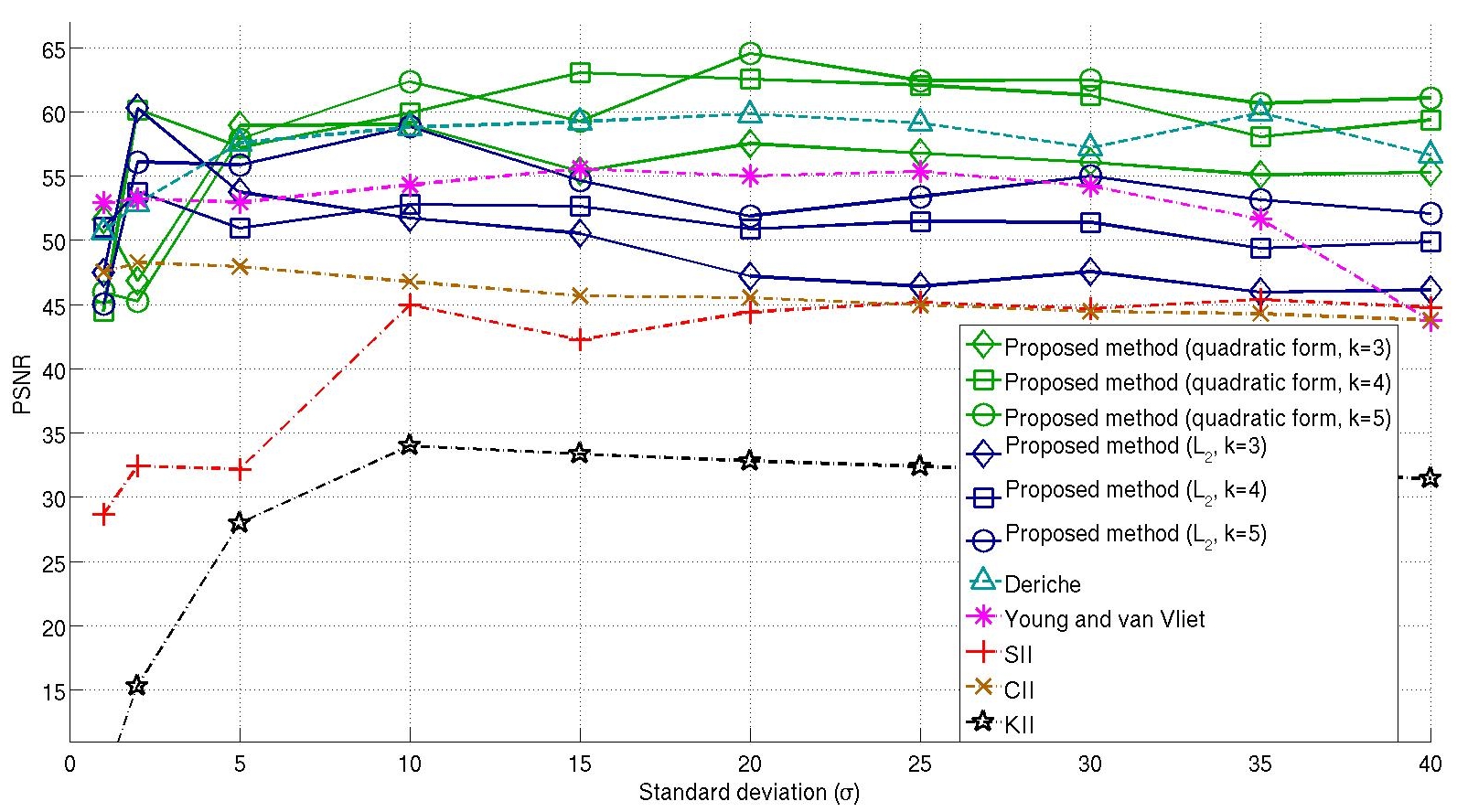}
  \label{fig:psnr}
  }


  \caption{
    Experimental results (Section~\ref{sec:res}).
    (a) Time speedup is in comparison to exact filtering. 
    Our proposed method with $k=3$ is the fastest.
    Using $k=5$ is similar to Young and van Vliet, while Deriche and SII are slower.
    (b) Accuracy of the output image:
    the highest PSNR (minimal $l_2$) is achieved by our proposed quadratic form
    approximation with $k=4$ or $k=5$ which are slightly better than Deriche.
    Using $k=3$ is still better than all other methods including Young and van Vliet.
    Approximating the Gaussian kernel by $l_2$ instead of the proposed quadratic form
    decreases the approximation accuracy.
    Kernel Integral Images (KII) gives poor results for small $\sigma$ values
    since integral images of polynomials are numerically unstable for
    high ratio between $\sigma$ and the image size.
  }
  \label{fig:res}
\end{figure*}

As shown by our experiments (Section~\ref{sec:res}, Figure~\ref{fig:psnr}), 
approximating the Gaussian kernel
using the proposed quadratic form (Equation~\ref{eq:mahal}) where $A_{jk} = \Phi_{j-k}$ 
results in a very low $l_2$ error (or high PSNR) on the output image.
Modifying $\Phi$ values slightly decreases the accuracy.

\subsection{Gaussian Approximation}
\label{sec:gapprox}

The Gaussian kernel is zero mean and its only parameter is 
standard deviation $\sigma$.
Actually, all the Gaussian kernels are normalized and scaled versions 
of the standard kernel $\exp( \frac{t^2}{2} )$.
Therefore the approximation need to be computed only once for some $\sigma_{_0}$.
For other $\sigma$ values the pre-computed solution of $N(t)$ is simply rescaled.

We approximate the Gaussian within the range $[-\pi\sigma,\pi\sigma]$,
since for greater distances from the origin kernel values are negligible.
The optimization is done on the positive part $[0,\pi\sigma]$.
Then the kernel symmetry is used to getcompute the weights $w_i$ as described 
in Section~\ref{subsec:sym}.

The partition indices $p_i$ and constants $c_i$ were found by an exhaustive search
using $100$ samples of the Gaussian kernel with $\sigma_{_0} = \frac{100}{\pi}$.
To get the parameters for other $\sigma$ values, we scale the indices and round them:

\begin{equation}
\label{eq:scale1}
p_i^{(\sigma)} = \left\lfloor \frac{\sigma}{\sigma_{_0}} \cdot p_i \right\rfloor
\end{equation}

\noindent
The weights are also scaled,

\begin{equation}
\label{eq:scale2}
w_i^{(\sigma)} = \frac{p_i}{2p_i^{(\sigma)}+1} \cdot w_i
\end{equation}

The parameters  for different numbers of constants
are presented in Table~\ref{tbl:params}.
Figure~\ref{fig:1Dapprox} shows the optimal approximation with $4$ constant functions.
Figure~\ref{fig:pulse}(c,d) present the 2D result of filtering image rows
and columns using the proposed approximation.

\begin{table}[h]

\begin{center}
\begin{tabular}{|c|l|l|}
\hline
\#constants ($k$) &  partition indices ($p_i$) &  weights ($c_i$)  \\
\hline
\hline
$k = 3$    &  23, 46, 76               & 0.9495, 0.5502, 0.1618\\
\hline
$k = 4$    &  19, 37, 56, 82           & 0.9649, 0.6700, 0.3376, 0.0976\\
\hline
$k = 5$    &  16, 30, 44, 61, 85       & 0.9738, 0.7596, 0.5031, \\
           &                           & 0.2534, 0.0739\\
\hline
\end{tabular}
\end{center}
\caption{Algorithm parameters.}
\label{tbl:params}

\end{table}

\section{Experimental Results}
\label{sec:res}

In order to evaluate the performance Gaussian filtering methods we made an experiment
on natural images from different scenes.
We used a collection of $20$ high resolution images (at least $1$ megapixel) taken from
Flickr under creative commons.
Each of the image was filtered with several standard deviation ($\sigma$) values.
Speed and accuracy are measured in comparison to the 
exact Gaussian filtering performed by the 'cvSmooth' function of the OpenCV library~\cite{opencv}.
The output image error is measured by the peak signal-to-noise ratio (PSNR).
This score is defined as $-10log_{10}(MSE)$, $MSE$ is the mean squared error 
and the image values are in $[0,1]$.

We examined our proposed method as well as 
the state of the art methods of Deriche~\cite{deriche1993recursively}
and Young and van Vliet~\cite{young1995recursive}.
We also examined the integral image based methods of
Hussein et al. (KII,~\cite{hussein2008kernel}),
Elboher and Werman (CII,~\cite{elboher2011cosine})
and Bhatia et al. (SII,~\cite{bhatia2010stacked}),
which use a similar approach to the proposed method (Section~\ref{sec:related}).
All the compared methods were implemented by us\footnote{
The CImg library contains an implemetation of  Deriche's method for $k=2$
coefficients. 
Due to caching considerations, our implementation is about twice as fast.
For the integral image based methods there is no public implementation
except CII~\cite{elboher2011cosine} which was proposed by the authors of the current paper.
} 
except Young and van Vliet's filtering, for which we 
adopted the code of~\cite{geusebroek2003fast}. 
This implementation is based on the updated design of Young and van Vliet 
in~\cite{young2002recursive} 
with corrected boundary conditions~\cite{triggs2006boundary}.

In order to examine the effect of using different error functions for 
kernel approximation, we tested our proposed method with two sets of
parameters.
The first set was computed by minimizing the  quadratic form error
function defined in Section~\ref{sec:l2}.
The second set was computed  by minimizing the $l_2$ error.
Since the filtering technique is identical the speed is the same, 
the  difference is
in the output accuracy.

Figure~\ref{fig:res} presents the average results of our experiments.
The highest acceleration is achieved by our proposed method with $k=3$ constants 
(Figure~\ref{fig:speedup}).
Using $k=4$ is still faster than all other methods, while $k=5$ has equivalent
speedup as Young and van Vliet's method.

The best accuracy is achieved by our quadratic form approximation with $k=4,5$,
which is slightly better than Deriche's method (Figure~\ref{fig:psnr}).
Using $l_2$ kernel approximation decreases the quality of our proposed method.
This demonstrates the importance of using an appropriate kernel approximation.
However, these results are still similar to Young and van Vliet and better than
other integral image based methods.

\section{Conclusion}
\label{sec:conc}

We present a very efficient and simple
scheme for filtering with separable non uniform kernels.
In addition we analyze the relation between kernel approximation, 
output error and natural image statistics.
Computing an appropriate Gaussian approximation by the proposed filtering scheme
is faster than the current state of art methods, while preserving similar accuracy.


%

\bibliographystyle{IEEEbib}

\bibliography{refs}

\end{document}